\documentclass[a4paper, 12pt]{article}

\usepackage[utf8]{inputenc}

\usepackage{amsfonts}
\usepackage{amssymb}
\usepackage{amsmath}
\usepackage{amsthm}

\usepackage{bbold}
\usepackage{bm}
\usepackage{graphicx}
\usepackage{color}
\usepackage{lscape}
\usepackage{makecell}
\usepackage{dirtytalk}
\usepackage{textcomp}
\usepackage{hyperref}

\usepackage{authblk}

\begin{document}

\title{Investigating the Monte-Carlo Tree Search Approach
for the Job Shop Scheduling Problem}

\author[1]{Boveroux Laurie}
\author[1]{Ernst Damien}
\author[1]{Louveaux Quentin}

\affil[1]{University of Liege}

%\vspace{1cm}
% \author[1]{Laurie Boveroux\corref{cor1}}
% \ead{laurie.boveroux@uliege.be}
% \affiliation[1]{organization={University of Liege},
% addressline={Allée de la Découverte 11},
% postcode={4000},
% city={Liege},
% country={Belgium}}

% \author[1]{Damien Ernst}
% \ead{dernst@uliege.be}

% \author[1]{Quentin Louveaux}
% \ead{q.louveaux@uliege.be}

% \cortext[cor1]{Corresponding author}

\begin{abstract}
The Job Shop Scheduling Problem (JSSP) is a well-known optimization problem in manufacturing, where the goal is to determine the optimal sequence of jobs across different machines to minimize a given objective.
In this work, we focus on minimising the weighted sum of job completion times.
We explore the potential of Monte Carlo Tree Search (MCTS), a heuristic-based reinforcement learning technique, to solve large-scale JSSPs, especially those with recirculation.
We propose several Markov Decision Process (MDP) formulations to model the JSSP for the MCTS algorithm.
In addition, we introduce a new synthetic benchmark derived from real manufacturing data, which captures the complexity of large, non-rectangular instances often encountered in practice.
Our experimental results show that MCTS effectively produces good-quality solutions for large-scale JSSP instances, outperforming our constraint programming approach.
\end{abstract}

% \begin{keyword}
% Job Shop Scheduling \sep Monte-Carlo Tree Search
% \end{keyword}

\maketitle 

\section{Introduction}
The Job Shop Scheduling Problem (JSSP) is a complex challenge faced by manufacturers.
The JSSP involves determining the optimal sequence of jobs on different machines to ensure that production processes are carried out efficiently.
This problem is important because it directly impacts a company's productivity, operating costs and ability to meet delivery schedules.
For example, delays in the production schedule can cause bottlenecks, higher inventory costs, and missed deadlines.
These issues can lead to unhappy customers and financial penalties.
Therefore, optimizing job scheduling is a practical need and a decisive strategy for success.

Many mathematical programming-based approaches exist to solve the JSSP, such as mixed-integer linear programming and constraint programming.
These methods are exact as they find optimal solutions by exhaustively exploring the search space.

However, they have limitations in practice.
The JSSP is known to be NP-hard.
As the number of jobs and machines increases, the complexity of the problem grows exponentially.
In practice, scheduling problems are often large-scale, dynamic and imbalanced.
In such scenarios, some machines may be heavily loaded while others are idle and processing times of the tasks can vary widely from a few units of time to several hundreds units.
In large-scale environments, exact methods often become impractical.

To address these limitations, approximate solutions have been developed.
Commonly used heuristics are Priority Dispatching Rules (PDRs) \cite{pinedo2012modeling}.
PDRs are simple heuristics that select the next operation to be scheduled based on a specific criterion.
These rules are easy to implement and computationally efficient, but they often offer low-quality solutions.
Another well-known heuristic approach is the Shifting Bottleneck Heuristic \cite{adams1988shifting}.
This procedure solves the problem by iteratively identifying the machine that creates a bottleneck and solving the schedule optimally on that unique machine using the one-machine schedule method by Carlier \cite{carlier1982one}.

Different meta-heuristic approaches such as simulated annealing, tabu search and genetic algorithms have been used to produce good quality solutions for the JSSP.
Simulated annealing \cite{Kirkpatrick1983}, an optimisation algorithm inspired by the process of physical annealing \cite{Rutenbar1989}, starts with an initial solution and iteratively refines it by making small random changes.
The algorithm evaluates the modified solution and accepts improvements.
However, it can also accept worse solutions with a probability that decreases over time, allowing the algorithm to escape local optima and explore a wider search space.
Tabu search \cite{Dell1993, Nowicki1996, Nowicki2005} also follows an iterative framework, but uses a deterministic acceptance-rejection criterion.
The key of this heuristic is a \say{tabu list} that prevents the algorithm from revisiting previously explored states or undoing recent modifications.
By searching the neighbourhood of the current solution and occasionally accepting worse schedules, the algorithm explores different solutions.
The tabu list ensures that unnecessary moves leading to revisited states are not allowed, so the algorithm can bypass local optima and explore potentially global solutions.

In recent years, learning-based methods have been developed.
Techniques such as deep reinforcement learning have been applied to learn scheduling policies.
For example, Zhang et al. \cite{zhang2020learning} proposed a graph neural network model with an actor-critic algorithm to learn effective dispatching rules.

% 6. Introduce your new one and mention why you believe it is promising.
One promising approach is the Monte Carlo Tree Search (MCTS) algorithm.
MCTS is a heuristic search algorithm that combines tree search with random sampling to find solutions in large search spaces.
Initially applied to the game domain, this reinforcement learning (RL) algorithm has shown its effectiveness in finding good strategies for complex games such as chess and Go \cite{silver2016mastering}.
The JSSP shares similarities with these games, as both involve making sequential decisions and optimizing outcomes based on a series of actions.
More specifically, we note that MCTS algorithm as been applied successfully to solve Markov Decision, which suggests their applicability to JSSP as they can be effectively framed as a Markov Decision Process (MDP).
States, actions, transition probabilities and rewards define an MDP.
Different MDP formulations can be employed to represent the scheduling problems.

Existing benchmarks, such as those proposed by Taillard et al. \cite{taillard1993benchmarks} and Adams et al. \cite{adams1988shifting}, are commonly used for testing new algorithms. 
However, these benchmarks are designed for smaller and simplified cases.
Such benchmarks do not capture the complexities of real-world scenarios, where machine loads can vary significantly, with some machines heavily loaded while others are idle.
These benchmarks are less suitable for evaluating algorithm performance in large-scale industrial scheduling problems.
To address this gap, we introduce a new synthetic benchmark derived from real-world manufacturing data that captures the complexity typical of large-scale industrial environments.
In this work, we investigate the potential of MCTS in solving the real large-scale JSSP.
The main contributions of this work are:
\begin{enumerate}
    \item We investigate and evaluate different ways to model JSSP as an MDP for the MCTS algorithm.
    \item We deliver a new benchmark created from anonymised real-world data \footnote{\url{https://github.com/LaurieBvrx/large-scale-complex-JSSP-benchmark.git}}.
\end{enumerate}

The rest of the paper is organized as follows.
In Section 2, we formally define the JSSP.
Section 3 describes our approach, including how we model JSSP as an MDP for MCTS and the constraint programming model used for comparison.
In Section 4, we explain the process of generating a new benchmark from anonymized real-world data.
Section 5 presents the experimental setup and results. Finally, Section 6 concludes the paper and outlines directions for future research.

\section{Problem Statement}
We consider the JSSP.
In a general JSSP, we are given a set $\mathcal{J}$ of $n$ jobs $J_1, J_2, ..., J_n$ and a set $\mathcal{M}$ of $m$ machines. 
Each job $J_i \in \mathcal{J}$ has an operation set $\mathcal{O}_i$ which contains $n_i$ operations $O_{ij}$ that must be processed in a specific order (i.e., with precedence constraints).
Each operation $O_{ij}$ of job $J_i$ requires a processing time $p_{ij}$ on a specific machine $M_{(ij)}$.
A job can have several operations that must be processed on the same machine (i.e., recirculation).
Each machine can process at most one operation at a time with no preemption.

To solve the JSSP, we must find a schedule that determines the order in which the operations are processed to minimize a specific objective function.
The objective most commonly minimized in the literature is the makespan, the maximum completion time of all operations in the schedule.
However, this objective does not consider the schedule's internal structure.
As we look only at the last operation to finish, we can open all jobs from the beginning.
Some jobs may be completed very late, making the open jobs wait for a long time and taking space in the inventory.
An objective that seems more realistic for real industrial problems is the weighted sum of the completion times of the $n$ jobs.
This objective is in line with the companies' objective of maximizing billings over time and should limit ongoing activities.
It can be formulated as follows:
\begin{equation*}
    \text{minimize} \quad \sum_{j \in \mathcal{J}} w_j \  C_j
\end{equation*}
where $w_j$ is the weight of the job $j$ and $C_j$ its completion time.

To describe a scheduling problem accurately, the standard notation in literature is the triplet $\alpha|\beta|\gamma$ where $\alpha$ represents the machine environment, $\beta$ the processing characteristics and the constraints and $\gamma$ the objective.
The problem we consider can be characterized by the triplet
\begin{equation*}
    J_m | prec, rcrc | w_j C_j
\end{equation*}
This triplet refers to a job shop environment with $m$ machines ($J_m$).
There are precedence constraints ($prec$), meaning that there are certain jobs or operations that must be completed before others can begin.
The other processing characteristic is the recirculation ($rcrc$), which implies that two or more operations of a job can be processed on the same machine.
In contrast to a classic job shop, where each job has exactly one operation on each machine, recirculation allows jobs to visit a machine more than once.
Finally, the objective is the minimisation of the total weighted completion times.

\section{Approach}
This section presents different approaches to solving the JSSP using MCTS.
First, we detail the basic concepts of a MDP.
Afterwards, we discuss how MCTS algorithms can be used to solve MDPs.
And, finally, we show various ways for casting the JSSP introduced in Section 3.3 as an MDP.

Recent studies have demonstrated the potential of MCTS in solving scheduling problems.
For instance, Saqlain, Ali, and Lee \cite{saqlain2023monte} proposed an MCTS-based algorithm for the flexible JSSP to minimize makespan.
Similarly, Chou et al. \cite{chou2015new} developed an approach that minimizes a multi-objective function using MCTS.
Building on this prior work, we propose additional MDP frameworks tailored to JSSPs and demonstrate how they can be effectively solved using MCTS methods.

\subsection{Markov Decision Process}
A Markov Decision Process (MDP) is a mathematical framework for modelling decision-making problems.
It is defined through the following objects: a
state space $\mathcal{S}$, an action space $\mathcal{A}$, transition probabilities $p(s'|s,a) \forall s, s' \in \mathcal{S}, a \in \mathcal{A}$ and a reward function $r(s, a)$.
The function $p(s'|s,a)$ gives the probability of reaching a state $s'$ after taking the action $a$ while being in state $s$.
At each time step $t$, the decision-maker observes the current state $s_t$ and selects an action $u_t$, which influences both the immediate reward and the state transition.

\subsection{Monte-Carlo Tree Search}

\begin{figure}
    \centering
    \includegraphics[width=1.0\linewidth]{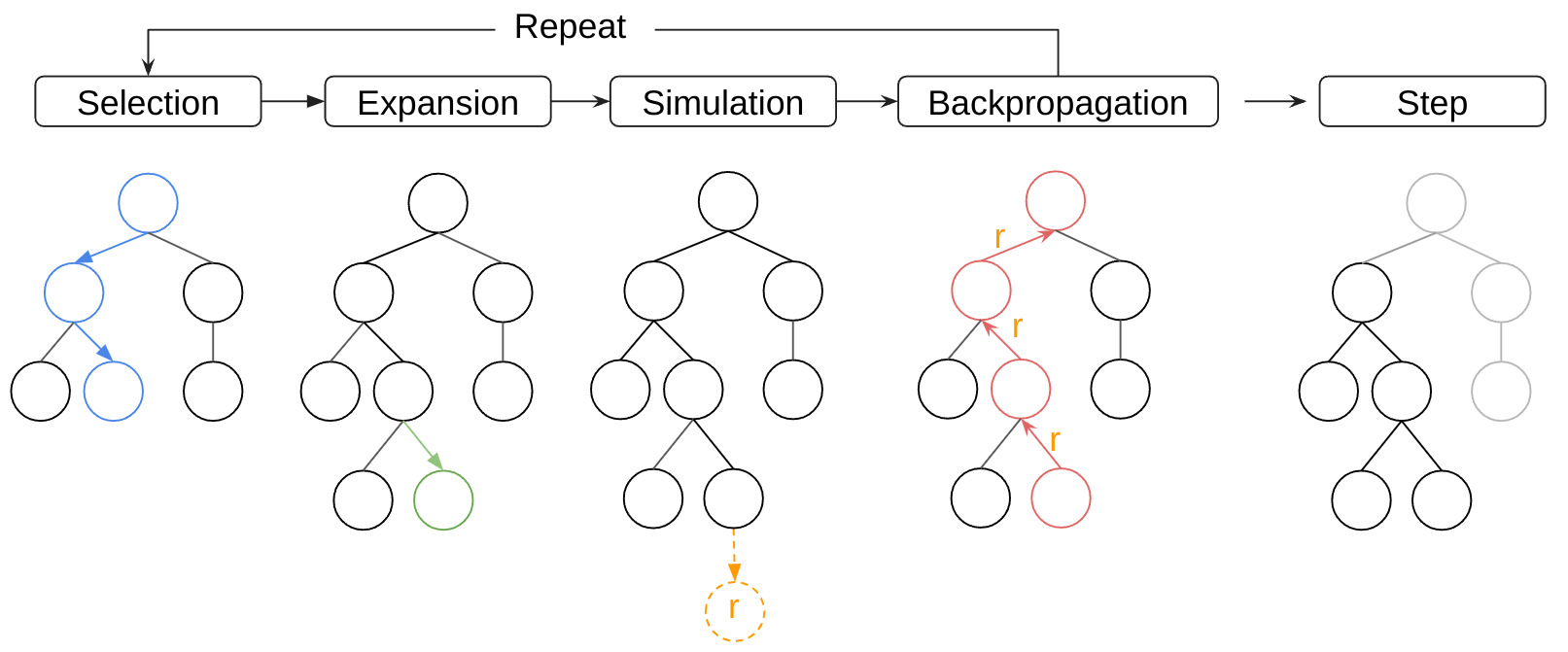}
    \caption{The four fundamental steps of the MCTS algorithm and the additional step.}
    \label{fig:mctssteps}
\end{figure}

Monte Carlo Tree Search (MCTS) is a heuristic search algorithm used in decision processes \cite{browne2012survey}.
It combines classic tree search implementations with machine learning principles of reinforcement learning to balance exploration and exploitation.
The algorithm is based on the building of a search tree.
The MCTS algorithm can be used to solve MDPs as it incrementally builds a search tree representing the states and actions of the MDP, using simulation-based techniques to evaluate potential policies. 
Indeed, each node of the tree represents a state of the decision process and each edge represents an action leading to a new state.
This algorithm aims to determine an optimal policy, i.e., a mapping from states to actions, that maximizes the expected cumulative reward over time, often defined as a discounted sum of rewards.
The algorithm is composed of four fundamental steps: selection, expansion, simulation and backpropagation.
These are schematically represented in Figure \ref{fig:mctssteps} \cite{browne2012survey}.

In the \textit{selection phase}, the algorithm starts from the root node at time 0 and from a node at deep $t$ at time $t$ that corresponds to the state $s_t$.
It successively selects a child node following a tree policy until it reaches a node that is not fully expanded.
A tree policy that has promising properties is the Upper Confidence Bound (UCB) formula.
This formula balances the exploitation of the best-known nodes and the exploration of less visited nodes.
A child node $j$ is selected based on the UCB formula:
\begin{equation}
    UCB = \bar{X}_j + C \sqrt{\frac{\ln N}{n_j}}
\end{equation}
The first part of the equation is the exploitation part, where $\bar{X}_j$ is the average reward of the node, and must be in $[0, 1]$.
The higher the average reward, the more the node is exploited.
The second part of the equation is the exploration part.
The total number of visits of the parent node is $N$ and the number of visits of the child node is $n_j$.
The lower the number of visits, the more the node is explored.
$C$ is a constant that controls the exploration-exploitation trade-off.
It is usually tuned empirically.
Note that if several child nodes have the same UCB value, one is selected randomly.

During the \textit{expansion phase}, a child node is added to the selected node according to the available actions.

In the \textit{simulation phase}, the algorithm extends nodes until a terminal state is reached following a default policy.
The default policy can be random or based on a heuristic.
In large-scale problems, the simulation step can be computationally expensive.
To reduce the computational cost, the simulation step follows a random policy.
Several simulations are performed from the current state to obtain a robust estimate of the value of the state, exploiting randomness to cover a wide range of possible outcomes.
%To reduce the computational cost, the simulation step can be replaced by a heuristic that estimates the value of the state.

Finally, in the \textit{backpropagation phase}, the algorithm updates the statistics of the nodes selected during the expansion and simulation steps.
The statistics are updated by backpropagating the evaluation of the simulation from the leaf node to the root node.
The evaluation is based on the reward function of the problem.

The algorithm repeats these steps until a stopping criterion is met.
The stopping criterion can be a fixed number of iterations or a time limit.

An additional step \textit{step} can be added to the MCTS algorithm to improve the search space exploration.
This step consists of selecting the best action to take at the current depth (starting from the root) and continuing the search from this node, i.e. from a deeper node in the tree.
This step is added after a certain number of repetitions when the node at the current depth is well explored.

\subsection{Modelling JSSP as an MDP}
This article explores different environments for the MCTS applied to the JSSP.
Each environment is defined by its state space, action space and reward function.
These three components can be defined in multiple ways, each influencing the performance and behaviour of the MCTS algorithm differently.
We explore a few of possible combinations of these components in the following.

\subsubsection*{State Space}
In reinforcement learning, the state $s_t$ is a representation of the situation of the agent at the decision step $t$.
In the context of JSSP, the state $s_t$ corresponds to the partial schedule of jobs at decision step $t$.
Here, $t$ in the MDP refers to the sequence of decision steps during the schedule construction and is unrelated to the actual physical timeline of the schedule.
We introduce two distinct state representations for the partial schedule:
\begin{enumerate}
    \item Absolute Representation: This representation maintains the completion times of each operation, directly encoding the timing information of the partial schedule.
    This approach is greedy regarding the order of operations and the completion times.
    \item Relative Representation: This representation maintains the order of the operations on each machine, encoding the sequence of operations on each machine rather than their precise timing.
    This approach is greedy only on the order of operations and it maintains flexibility in terms of scheduling completion time.
\end{enumerate}
We can readily convert the representation of one type to the other.
These two representations are convertible.
Starting from the relative representation, we can compute the completion times $C_j$ for all operations, provided the order of operations on every machine is known.
This computation can be achieved in $\mathcal{O}(n\log n)$ where $n$ is the total number of operations.
On the other hand, we can derive the relative representation from the absolute one by ordering the operations on each machine based on their completion times.

% \textbf{State representation 1: }
% The state $s_t$ at time $t$ can be represented by two components:
% \begin{enumerate}
%     \item A vector $\mathbf{C}_{t}=\left(C_{1}, C_{2}, \ldots, C_{n}\right)$, where $C_i$ denotes the completion time of the $i$-th operation that has been scheduled.
%     \item A set $\mathcal{\bar O}_{t}$ of operations that have not yet been scheduled.
% \end{enumerate}
% Thus the state $s_t$ can be expressed as:
% \begin{equation*}
%     s_{t}=\left(\mathbf{C}_{t}, \mathcal{\bar O}_{t}\right)
% \end{equation*}

% \textbf{State representation 2: }
% Another approach is to define the state $s_t$ by:
% \begin{enumerate}
%     \item A sequence $\mathbf{O}_{m}=\left(o_{m 1}, o_{m 2}, \ldots, o_{m k}\right)$ for each machine $m \in\{1,2, \ldots, M\}$,
%     where each sequence represents the order of operations scheduled on machine $m$.
%     \item A set $\mathcal{\bar O}_{t}$ of operations that have not yet been scheduled.
% \end{enumerate}
% Let $\mathbf{O}_{t}=\left(\mathbf{O}_{1}, \mathbf{O}_{2}, \ldots, \mathbf{O}_{M}\right)$ be the collection of sequences for all machines.
% The state $s_t$ can then be expressed as:
% \begin{equation*}
%     s_{t}=\left(\mathbf{O}_{t}, \mathcal{\bar O}_{t}\right)
% \end{equation*}

\subsubsection*{Action Space}
The action space is the set of possible actions that the agent can take in a given state.
In the context of the JSSP, the action space can be defined by the selection of the operation(s) to be scheduled.
We propose four types of actions.

\begin{enumerate}
    \item The first approach is to select a single operation based on different dispatching rules (PDRs).
    PDRs are simple heuristics that select the next operation to be scheduled based on some criterion.
    For example, the first in first out (FIFO) rule schedules the next operation in order of appearance, the shortest processing time (SPT) rule selects the operation with the shortest processing time and the most operation remaining (MOR) rule selects the first operation available of the job with the most remaining operations.
    More details on the different PDRs used are given in Section \ref{sec:dispatching}.
    \begin{equation*}
        A=\{a \mid a=\operatorname{PDR}(\mathcal{\bar O}), \mathrm{PDR} \in\{\mathrm{FIFO}, \mathrm{SPT}, \mathrm{MOR}, \ldots\}\}
    \end{equation*}
    where $\mathcal{\bar O}$ is the set of operations that have not yet been scheduled, and PDR represents a dispatching priority rule.
    All operations in the selected job are scheduled in the order they appear.

    \item A second approach is to select an entire job based on a PDR.
    All operations in the selected job are scheduled in the order they appear in the job.
    \begin{equation*}
        A=\{a \mid a=\operatorname{PDR}({\mathcal{\bar J}}), \mathrm{PDR} \in\{\mathrm{FIFO}, \mathrm{SPT}, \mathrm{MOR}, \ldots\}\}
    \end{equation*}
    where $\mathcal{\bar J}$ is the set of jobs that have not yet been scheduled, and PDR represents a dispatching priority rule.
    All operations in the selected job are scheduled in the order they appear.
    Scheduling a whole job as a single action allows the search tree to be less deep, making the exploration space more manageable.

    \item A third approach is to select an operation based on a single PDR and then select a percentage.
    This percentage determines the size of the gap in the schedule of the corresponding machine in which the operation is scheduled.
    For example, if the percentage is 50\%, the operation will be scheduled in the first idle time in the schedule that is greater than 50\% of the operation's processing time.
    This action type is feasible when using the relative state representation, where the completion times of operations are not explicitly encoded and only the sequencing of operations is maintained.
    In this representation, even if there appears to be insufficient space theoretically, all subsequent operations can be shifted to accommodate the new operation, as the representation focuses sol, as the representation focuses only on the order of operations.
    Formally, let $A$ be the action space, then:
    \begin{equation*}
        A=\{a \mid a=(\operatorname{PDR}(\mathcal{\bar J}), p), \mathrm{PDR} \in\{\mathrm{FIFO}, \mathrm{SPT}, \mathrm{MOR}, \ldots\}, p \in[0,1]\}
    \end{equation*}
    where $\mathcal{\bar O}$ is the set of operations that have not yet been scheduled, PDR represents a dispatching priority rule.
    and $p$ is the percentage gap for the scheduling of the selected operation in the corresponding machine's schedule.

    \item The fourth approach is similar to the third one except that we schedule all the operations of the selected job in one step.
    %A fourth approach is to select a job based on a single PDR and then select a percentage.
    %This percentage is fixed for each operation of the selected job.
    %All operations in the selected job are scheduled in the order they appear in the job.
    \begin{equation*}
        A=\{a \mid a=(\operatorname{PDR}(\mathcal{\bar J}), p), \mathrm{PDR} \in\{\mathrm{FIFO}, \mathrm{SPT}, \mathrm{MOR}, \ldots\}, p \in[0,1]\}
    \end{equation*}
    where $\mathcal{\bar J}$ is the set of jobs that have not yet been scheduled, PDR represents a dispatching priority rule.
    
    \end{enumerate}
%\textbf{Type of action 1: }
%\textbf{Type of action 2: }
%\textbf{Type of action 3: }
%\textbf{Type of action 4: }

\subsubsection*{Reward}
The reward function is related to the objective function of the problem and is designed to reflect the quality of the solution.
The most intuitive way to define the reward function in the context of the JSSP would be to set the reward to 0 unless a terminal state has been reached.
A terminal state is a state reached when the problem is solved (i.e., all jobs are completed) or it is impossible to continue (for example, due to infeasibility, such as exceeding resource limits or invalid machine assignments).
In such cases, the reward is defined differently:
\begin{itemize}
    \item If the terminal state represents a successfully completed schedule, the reward is the negative weighted sum of the completion times of all jobs, aligning with the objective to minimize the total weighted completion time.
    \item If the terminal state represents an infeasible solution, the reward is $-\infty$.
\end{itemize}

The reward can also be normalised between 0 and 1:
\begin{equation*}
    \hat{r} = \frac{r - r_{\text{min}}}{r_{\text{max}} - r_{\text{min}}}
\end{equation*}
where $r_{\text{min}}$ and $r_{\text{max}}$ are the minimum and maximum possible rewards, respectively.
%$r_{\text{min}}$ is given by the lower bound of the problem expressed with a linear program, and $r_{\text{max}}$ is computed by the sum of the processing times of all operations.
The variable $r_{\text{min}}$ represents the total processing time of all operations, corresponding to a scenario where all jobs are processed in parallel without any restrictions.
On the other hand, $r_{\text{max}}$ is defined as the sum of the completion times when jobs are processed sequentially.
In this case, the completion time of a job $i$ is the completion time of the previous job $i-1$ plus the total processing time of all operations of job $i$.
Normalisation is necessary to ensure the UCB formula is in the range $[0,1]$.

\subsection{Priority Dispatching Rules}
\label{sec:dispatching}
PDRs are simple heuristics that select the next operation or job to be scheduled based on a specific criterion \cite{reijnen2023job}.
They are widely used in practice because they are easy to implement and computationally efficient, particularly for large-scale problems.
They can either focus on entire jobs or individual operations.
At a job-level selection, a PDR prioritizes jobs based on the characteristics of different jobs.
On the other hand, at the operation-level selection, a PDR prioritizes an individual operation.
The operation can be selected based on their own characteristics (e.g., processing time) or because they are the first available operation of the job chosen by a job-level selection.

The list of the job-level PDRs are listed in the following:
\begin{itemize}
    \item The FIFO rule (first in first out) processes jobs in the order they are given in the instance.\\
           If $J_i$ and $J_j$ are two jobs, then under FIFO:
            \begin{equation*}
                \text { If } i<j \text {, then } J_{i} \text { is processed before } J_{j} \text {. }
            \end{equation*}

    \item The Least Work First (LWF) rule selects the job with the shortest total processing time.
        The processing time of a job is the sum of the processing times of all its operations.
        If $J_i$ and $J_j$ are two jobs and $P_i$ represents the total processing time for job $J_i$, then under LWF:
        \begin{equation*}
            \text{ If } P_i < P_j \text{, then } J_i \text{ is scheduled before } J_j \text {. }
        \end{equation*}
        
    \item The Most Work First (MWF) rule selects the job with the longest total processing time.

    \item The Shortest Job First (SJF) rule prioritizes jobs with the least number of operations.\\
        If $J_i$ and $J_j$ are two jobs and $n_i$ represents the number of operations of job $J_i$, then under SJF:
        \begin{equation*}
            \text{ If } n_i < n_j \text{, then } J_i \text{ is scheduled before } J_j \text {. }
        \end{equation*}

    \item The Largest Job First (LJF) rule selects the job with the largest number of operations.
\end{itemize}

The list of the operation-level PDRs are listed in the following:
\begin{itemize}
    \item The FIFO rule (first in first out) processes operations in the order they are given in the instance.\\
           If $O_i$ and $O_j$ are two operations, then under FIFO:
            \begin{equation*}
                \text { If } i<j \text {, then } O_{i} \text { is processed before } O_{j} \text {. }
            \end{equation*}

    \item The Least Work Remaining (LWR) rule prioritizes the first available operation of the job with the least total processing time remaining across all jobs.
        If $J_i$ and $J_j$ are two jobs, $O_i$ and $O_j$ are the first available operations of jobs $J_i$ and $J_j$ and $P_{i,t}$ represents the total processing time remaining for job $J_i$ at step $t$, then under LWR:
        \begin{equation*}
            \text{ If } P_{i,t} < P_{j,t} \text{, then } O_i \text{ is scheduled before } O_j \text {. }
        \end{equation*} 
        
    \item The Most Work Remaining (MWR) rule is similar to the LWR rule, but instead of prioritizing the least total processing time, it prioritises the first available operation of the job with the most total processing time remaining.
    
    \item The Least Operations Remaining (LOR) rule prioritizes the first available operation of the job with the least number of operations remaining across all jobs.\\
        If $J_i$ and $J_j$ are two jobs, $O_i$ and $O_j$ are the first available operations of jobs $J_i$ and $J_j$ respectively and $n_{i, t}$ represents the number of operations that still have to be scheduled for job $J_i$ at step $t$, then under LOR:
        \begin{equation*}
            \text{ If } n_{i,t} < n_{j,t} \text{, then } O_i \text{ is scheduled before } O_j \text {. }
        \end{equation*}

    \item The Most Operations Remaining (MOR) rule is similar to the LOR rule, but instead of prioritizing the least number of operations, it prioritizes the first available operation of the job with the most operations remaining.

    \item The Shortest Processing Time (SPT) rule selects the operation with the shortest processing time.
        If $O_i$ and $O_j$ are two operations and $p_i$ and $p_j$ are their processing times, then under SPT:
        \begin{equation*}
            \text{ If } p_i < p_j \text{, then } O_i \text{ is processed before } O_j \text {. }
        \end{equation*} 
        
    \item The Longest Processing Time (LPT) rule selects the operation with the longest processing time.    
\end{itemize}

\subsection{Constraint programming}
\label{sec:constraint}
Constraint programming is a declarative programming paradigm for modelling and solving combinatorial problems.
By integrating constraint programming into our analysis, we can compare its performance with the MCTS approach.
This method is more sophisticated and involves a longer computational process than PDRs.
As MCTS typically requires considerable computing time to converge on good solutions, the use of constraint programming allows a fairer comparison of results.

The constraint programming model we use is defined in the following:

\subsubsection*{Variables}
    \begin{itemize}
        \item $start_i$: Start time of operation $i$, where $i = 1, \ldots, n$ (where $n$ is the number of operations).
        \item $end_i$: End time of operation $i$, where $i = 1, \ldots, n$.
    \end{itemize}
    
\subsubsection*{Data}
    \begin{itemize}
        \item $duration_i$: Processing time of operation $i$.
        \item $machine_i$: Machine assigned to operation $i$.
    \end{itemize}
    
\subsubsection*{Objective}
    Minimize the total completion time of jobs, i.e. of the last operations of each job:
        \[\min \sum_{i \in \text{terminal\_operations}} end_i   \]
        
\subsubsection*{Constraints}
    \begin{itemize}
    \item Precedence constraints:
        For each operation $i$, and its predecessors $j \in \text{predecessors}[i]$:
        \[ end_i \geq end_j + duration_i    \]
        \item No-Overlap Constraints:
        For each machine $m$, ensure that two intervals do not overlap:
        \[ \text{NoOverlap}(\{(start_i, duration_i) \mid \text{machine\_i} = m\})   \]
    \end{itemize}

Along with the constraint programming model, we guide the search process by using the LWR PDR, which helps refine the search strategy and improve the efficiency of finding solutions.

\section {Data Generation}
There is a gap in the existing literature regarding job shop scheduling benchmarks.
Most commonly referenced instances, such as those proposed by Taillard et al. \cite{taillard1993benchmarks}, Adams et al. \cite{adams1988shifting} or Demirkol et al. \cite{demirkol1998benchmarks}, focus on small and rectangular configurations where the number of machines equals the number of operations of each job.
This structure does not adequately represent the complexities of larger, unbalanced scenarios commonly encountered in real-world manufacturing.

To address this gap, we analyze a job shop instance derived from a real-world manufacturing industry that includes 51 machines, 828 jobs and a total of 6057 operations.
In this instance, the workload distribution is unbalanced, with some machines heavily loaded while others are lightly used.
Furthermore, the number of operations per job varies significantly, ranging from 1 to 20.

To better simulate our real-world case, we generate a synthetic job shop scheduling benchmark based on the original instance.
The data generation process includes the creation of job and machine configurations that reflect the original conditions while including some level of variability and noise.
An overview of this process is detailed in the following.

\begin{enumerate}
    \item Job Configuration: we first generate a random integer between 600 and 1000 to determine the number of jobs.
    A type and a size are assigned to each job.
    We have two different types of jobs: common and unique.
    Common jobs are job types that occur more frequently in the job shop scheduling environment. They correspond to more frequent sets of pieces to manufacture.
    Unique jobs are job types that occur less frequently. They correspond to unique orders a manufacturing industry can receive.
    The sizes are drawn from a Gaussian distribution with a mean and standard deviation derived from the original instance.
    
    \item Machine configuration: we first generate a random integer between 50 and 70 to define the number of machines.
    They are then split into different types based on their operational characteristics.
    A specific distribution of the number of operations is assigned to each type.
    Once the machine types are identified, we introduce noise to the probability distribution of these types.
    The number of machines of each type is then determined by sampling from this noisy distribution and their number of operations are drawn.
    
    \item Operations to jobs assignment: with machine types and their distributions defined, we assign operations to jobs.
    Each job’s operations are distributed across the available machines based on the machine type distribution.
    We also ensure that the processing times for each operation are generated from a normal distribution centred around the mean processing time for the machine type, with a specified standard deviation.
    
\end{enumerate}
By following this process, we generated 20 instances of the JSSP based on the original data.
These instances are used to evaluate the performance of the MCTS algorithm with different scenarios.

\section{Experiments}
\subsection{Setup}
In this work, we evaluate the performance of five different types of environments for the MCTS algorithm.
Each of these types of environment has three possible actions in its action space, except for the fourth type, which has six actions.
The state representation, the type of action and the corresponding set of dispatching rules and, if needed, the percentages are listed in the Table \ref{tab:environment_summary}.

We note that we encountered significant computational problems due to the long-running time when evaluating the performance of the MCTS algorithm in environment type 3.
Specifically, the number of operations is high, making the search process computationally expensive.
One of the key issues arises from the need to recompute the completion times of all operations to identify the idle time at each new state.
This means that at each new scheduled operation, we recompute all the completion times.
As a result, this type of environment is not feasible for our use.

The MCTS algorithm is run for six repetition steps and 30 evaluations for the backpropagation phase.
The results are compared to the constraint programming model \ref*{sec:constraint}.
All the algorithms are coded in Python and the constraint programming model is solved using the Google OR-Tools library \cite{cpsatlp}.
The computations are executed on a calculation server with 48 Intel Xeon E7 v4 2.20 GHz processors with a total RAM of 128 GB.
%\raisebox{1ex}{\small{\textregistered}}

%\begin{landscape}
\begin{table}
    \centering
    \renewcommand{\arraystretch}{1.4}
    \begin{tabular}{|c|c|c|l|}
        \hline
        \textbf{\makecell{Env. \\ Type}} & \textbf{\makecell{State \\ Representation}} & \textbf{\makecell{Type of \\ Action}} & \textbf{$\mathrm{PDR}$ and $p$} \\
        \hline

        \textbf{Type 1.1} & absolute & 1 &   $  \{ \text{FIFO, LWR, MWR} \}$ \\
        \textbf{Type 1.2} & absolute & 1 &   $  \{ \text{FIFO, LOR, MOR} \}$ \\
        \textbf{Type 1.3} & absolute & 1 &   $  \{ \text{FIFO, SPT, LPT} \}$ \\
        \textbf{Type 1.4} & absolute & 1 &   $  \{ \text{LWR, LOR, SPT} \}$ \\
        \hline

        \textbf{Type 2.1}  \cite{saqlain2023monte} & absolute & 2 &   $  \{ \text{FIFO, SJF, LJF} \}$ \\
        \textbf{Type 2.2} & absolute & 2 &   $  \{ \text{FIFO, LWF, MWF} \}$ \\
        \textbf{Type 2.3} & absolute & 2 &   $  \{ \text{FIFO, SJF, LWF} \}$ \\
        \hline

        \textbf{Type 3} & relative & 3 &  \makecell{\textit{Not applicable due to}\\ \textit{high computational cost}} \\
        \hline
         
        \textbf{Type 4.1} & relative & 4 &   $  \{ \text{LWF}\}, [0.6, 0.8, 1.0]$ \\
        \textbf{Type 4.2} & relative & 4 &   $  \{ \text{LWF}\}, [0.3, 0.6, 0.8]$ \\
        \textbf{Type 4.3} & relative & 4 &   $  \{ \text{MWF}\}, [0.6, 0.8, 1.0]$ \\
        \textbf{Type 4.4} & relative & 4 &   $  \{ \text{MWF}\}, [0.3, 0.6, 0.8]$ \\
        \textbf{Type 4.5} & relative & 4 &   $  \{ \text{SJF}\}, [0.6, 0.8, 1.0]$ \\
        \textbf{Type 4.6} & relative & 4 &   $  \{ \text{SJF}\}, [0.3, 0.6, 0.8]$ \\
        \textbf{Type 4.7} & relative & 4 &   $  \{ \text{LJF}\}, [0.6, 0.8, 1.0]$ \\
        \textbf{Type 4.8} & relative & 4 &   $  \{ \text{LJF}\}, [0.3, 0.6, 0.8]$ \\
        \hline

        \textbf{Type 5.1} & relative & 4 &   $  \{ \text{LWF, MWF}\}, [0.6, 0.8, 1.0]$ \\
        \textbf{Type 5.2} & relative & 4 &   $  \{ \text{LWF, MWF}\}, [0.3, 0.6, 0.8]$ \\
        \textbf{Type 5.3} & relative & 4 &   $  \{ \text{SJF, LJF}\}, [0.6, 0.8, 1.0]$ \\
        \textbf{Type 5.4} & relative & 4 &   $  \{ \text{SJF, LJF}\}, [0.3, 0.6, 0.8]$ \\
        \textbf{Type 5.5} & relative & 4 &   $  \{ \text{LWF, SJF}\}, [0.6, 0.8, 1.0]$ \\
        \textbf{Type 5.6} & relative & 4 &   $  \{ \text{LWF, SJF}\}, [0.3, 0.6, 0.8]$ \\
        \hline
    \end{tabular}
    \caption{Summary of Environment Types and Characteristics for MCTS Algorithm Evaluation}
    \label{tab:environment_summary}
\end{table}
%\end{landscape}

\subsection{Results}
Using performance profiles, we first explore the different configurations with the same environment type and then compare the best configurations across all environment types.
A performance profile represents the percentage y of instances for which a specific method produces a solution whose objective function is not worse than x times the best solution found by any of the studied methods.

\begin{figure}   

    \centering
    \includegraphics[width=1.0\linewidth]{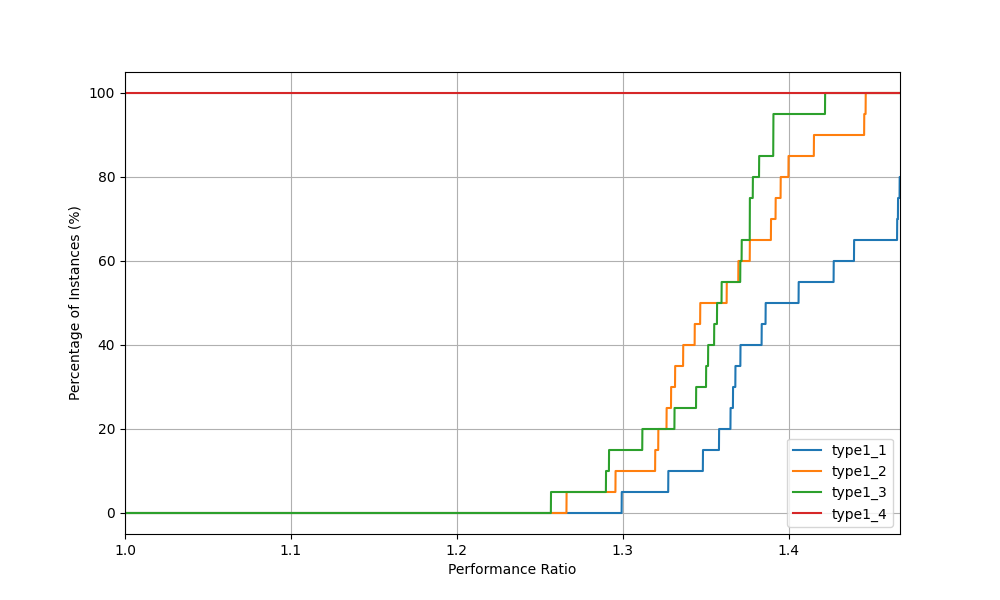}
    \caption{Performance profiles of the configurations of Type 1.}
    \label{fig:results1}
\end{figure}

\begin{figure}
    \centering
    \includegraphics[width=1.0\linewidth]{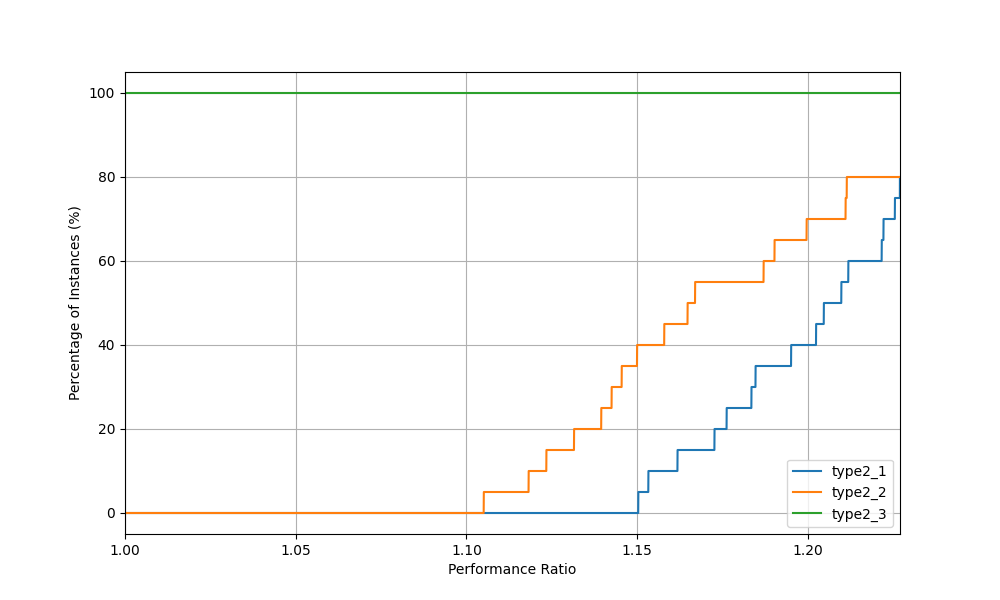}
    \caption{Performance profiles of the configurations of Type 2.}
    \label{fig:results2}
\end{figure}

\begin{figure}
    \centering
    \includegraphics[width=1.0\linewidth]{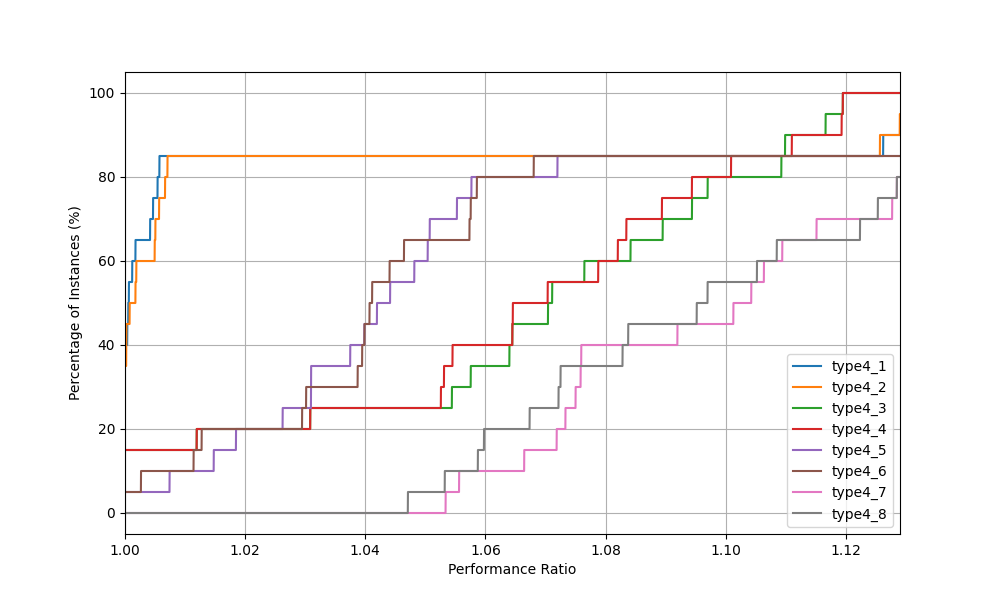}
    \caption{Performance profiles of configurations of Type 4.}
    \label{fig:results4}
\end{figure}

\begin{figure}
    \centering
    \includegraphics[width=1.0\linewidth]{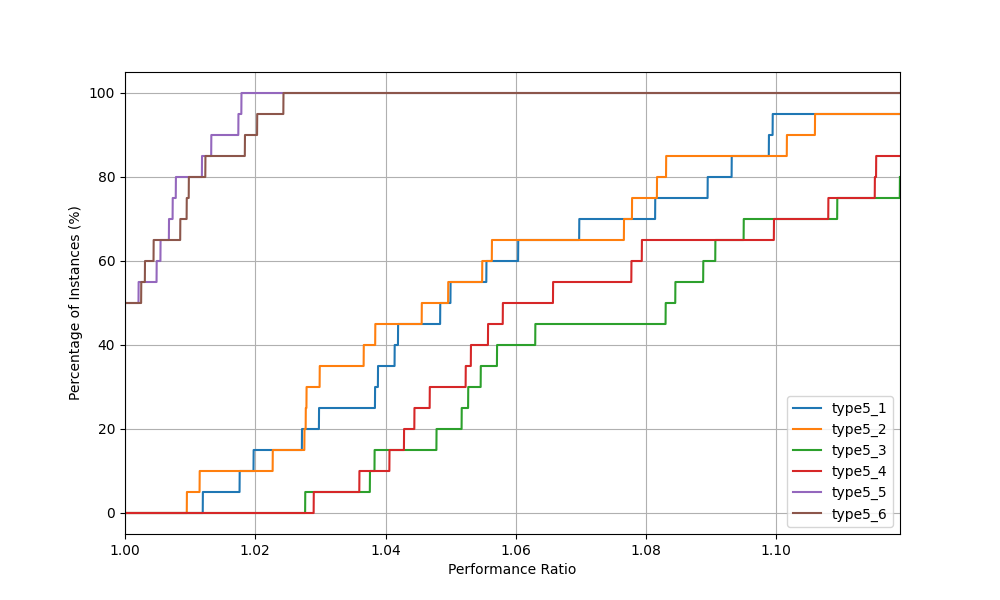}
    \caption{Performance profiles of the configurations of Type 5.}
    \label{fig:results5}
\end{figure}

Each figure compares all the variants of one specific type of environment outlined in Table \ref{tab:environment_summary}.
In Figure \ref{fig:results1}, the performance profiles indicate that Configuration 1.4 consistently outperforms all other configurations of Type 1.
Figure \ref{fig:results2} evaluates the configurations of Type 2.
Configuration 2.3 dominates the other configurations, achieving the best performance across all instances for the configuration Type 2.
Figure \ref{fig:results4} presents the results for configurations of Type 4.
Configurations 4.1 and 4.2 achieve the best performance, dominating the other configurations.
Configurations with the same PDR but different percentages have similar performance, indicating that the PDR is a key determining factor.
The results suggest that configurations using the LWF PDR consistently perform better, followed by those with SJF, MWF and finally LJF.
Figure \ref{fig:results5} compares configurations of Type 5, which demonstrate a similar pattern to Type 4.
Configurations 5.6 and 5.7 outperform the other configurations. Again, configurations with the same pair of PDRs but different percentages perform equivalently.
Configurations with LWF and SJF outperform those using LWF and MWF, which in turn perform better than those using SJF and LJF.

Additionally, the detailed results, including the mean performance across all instances for each configuration, are presented in Table \ref{tab:performance_summary}.

\begin{figure}
    \centering
    \includegraphics[width=1.0\linewidth]{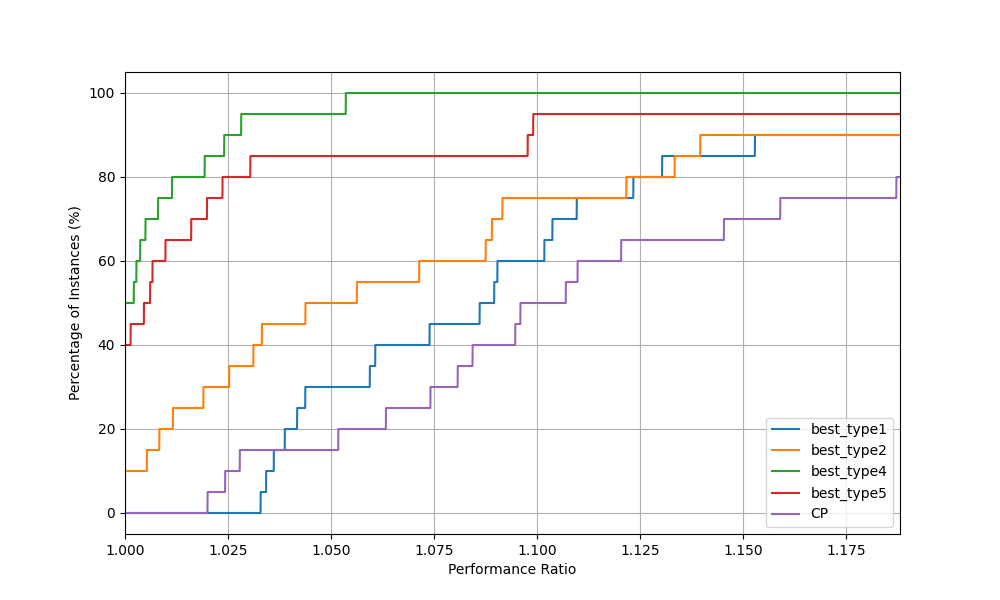}
    \caption{Performance profiles comparing the best-performing configuration of Types 1, 2, 4 and 5, along with the Constraint Programming results.}
    \label{fig:resultsAll}
\end{figure}

Finally, we compare the best-performing configurations of Types 1, 2, 4, and 5 alongside the constraint programming results.
Figure \ref{fig:resultsAll} gives the resulting performance profiles.
We observe that configurations from Types 4, 5 and 2 achieve the best performance for 50\%, 40\% and 10\% of the instances, respectively.
This highlights the benefit of using a less greedy and more flexible environment, as seen in Types 4 and 5, which process operations during idle time even if the operation processing time exceeds the available time.
Additionally, the results indicate that the MCTS-based algorithm outperforms the CP approach on the large instances we have considered, even when the latter is paired with a search heuristic.

\begin{table}
    \centering
    \begin{tabular}{|c|c|}
         \hline
         Method &  Mean \\
         \hline
         Constraint programming & 1.7136 \\
         \hline
         MCTS - Type 1.1 &  2.4014\\
         MCTS - Type 1.2 &  2.2852\\
         MCTS - Type 1.3 &  2.2859\\
         MCTS - Type 1.4 &  \textbf{1.6837}\\
         \hline
         MCTS - Type 2.1 &  2.0226\\
         MCTS - Type 2.2 &  1.9875\\
         MCTS - Type 2.3 &  \textbf{1.6792}\\
         \hline
         MCTS - Type 4.1 & \textbf{1.5532}\\
         MCTS - Type 4.2 &  1.5533\\
         MCTS - Type 4.3 &  1.5815\\
         MCTS - Type 4.4 &  1.5786\\
         MCTS - Type 4.5 &  1.6234\\
         MCTS - Type 4.6 &  1.6225\\
         MCTS - Type 4.7 & 1.6586 \\
         MCTS - Type 4.8 &  1.6552\\
         \hline
         MCTS - Type 5.1 &  1.6312\\
         MCTS - Type 5.2 &  1.6249\\
         MCTS - Type 5.3 &  1.6763\\
         MCTS - Type 5.4 &  1.6644\\
         MCTS - Type 5.5 &  \textbf{1.5415}\\
         MCTS - Type 5.6 &  1.5437\\
         \hline
    \end{tabular}
    \caption{Mean of the Weighted Sum of the Completion Times (scaled down by a factor of $10^8$) Across All Instances of Each Configuration of the MCTS and of the Constraint Programming Approach.}
    \label{tab:performance_summary}
\end{table}

\section{Conclusion}
In this study, we explore the potential of using MCTS to solve large-scale and real-world instances of the JSSP.
We introduced various MDP formulations to model the JSSP for the MCTS algorithm and compared their performance with a constraint programming model.
In addition, we deliver a new synthetic benchmark derived from anonymised real-world manufacturing data that captures the complexity and variability of industrial scheduling environments.
Our experimental results showed that MCTS is a promising approach for solving large-scale JSSPs, consistently outperforming our constraint programming approach.
The MCTS-based algorithm showed better performance in different MDP formulations.
In particular, configurations that allow operations to be inserted in idle time lower than their processing time are beneficial.
The proposed MDP formulations provide a flexible framework for representing different scheduling problems and the new benchmark is a valuable tool for testing and evaluating scheduling algorithms in industrial contexts.
Future research could further refine the MCTS approach by exploring a machine-learning-based reward function allowing the evaluation of a partial schedule.
In conclusion, our results support the utility of MCTS as an alternative heuristic to solve large job shop scheduling problems.

\bibliographystyle{unsrt}
\bibliography{bibliography.bib}

\begin{thebibliography}{10}

\bibitem{pinedo2012modeling}
Michael~L Pinedo.
\newblock Modeling and solving scheduling problems in practice.
\newblock {\em Scheduling: Theory, Algorithms, and Systems}, pages 431--458,
  2012.

\bibitem{adams1988shifting}
Joseph Adams, Egon Balas, and Daniel Zawack.
\newblock The shifting bottleneck procedure for job shop scheduling.
\newblock {\em Management science}, 34(3):391--401, 1988.

\bibitem{carlier1982one}
Jacques Carlier.
\newblock The one-machine sequencing problem.
\newblock {\em European Journal of Operational Research}, 11(1):42--47, 1982.

\bibitem{Kirkpatrick1983}
Scott Kirkpatrick, C.~Gelatt, and M.~Vecchi.
\newblock Optimization by simulated annealing.
\newblock {\em Science (New York, N.Y.)}, 220:671--80, 06 1983.

\bibitem{Rutenbar1989}
R.A. Rutenbar.
\newblock Simulated annealing algorithms: an overview.
\newblock {\em IEEE Circuits and Devices Magazine}, 5(1):19--26, 1989.

\bibitem{Dell1993}
Mauro Dell'Amico and Marco Trubian.
\newblock Applying tabu search to the job-shop scheduling problem.
\newblock {\em Annals of Operations Research}, 41:231--252, 09 1993.

\bibitem{Nowicki1996}
E.~Nowicki and C.~Smutnicki.
\newblock A fast taboo search algorithm for the job shop problem.
\newblock {\em Management Science}, 42(6):797--813, 1996.

\bibitem{Nowicki2005}
Eugeniusz Nowicki and Czesław Smutnicki.
\newblock An advanced tabu search algorithm for the job shop problem.
\newblock {\em Journal of Scheduling}, 8:145--159, 04 2005.

\bibitem{zhang2020learning}
Cong Zhang, Wen Song, Zhiguang Cao, Jie Zhang, Puay~Siew Tan, and Xu~Chi.
\newblock Learning to dispatch for job shop scheduling via deep reinforcement
  learning.
\newblock {\em Advances in neural information processing systems},
  33:1621--1632, 2020.

\bibitem{silver2016mastering}
David Silver, Aja Huang, Chris~J Maddison, Arthur Guez, Laurent Sifre, George
  Van Den~Driessche, Julian Schrittwieser, Ioannis Antonoglou, Veda
  Panneershelvam, Marc Lanctot, et~al.
\newblock Mastering the game of go with deep neural networks and tree search.
\newblock {\em nature}, 529(7587):484--489, 2016.

\bibitem{taillard1993benchmarks}
Eric Taillard.
\newblock Benchmarks for basic scheduling problems.
\newblock {\em european journal of operational research}, 64(2):278--285, 1993.

\bibitem{saqlain2023monte}
M~Saqlain, S~Ali, and JY~Lee.
\newblock A monte-carlo tree search algorithm for the flexible job-shop
  scheduling in manufacturing systems.
\newblock {\em Flexible Services and Manufacturing Journal}, 35(2):548--571,
  2023.

\bibitem{chou2015new}
Jen-Jai Chou, Chao-Chin Liang, Hung-Chun Wu, I-Chen Wu, and Tung-Ying Wu.
\newblock A new mcts-based algorithm for multi-objective flexible job shop
  scheduling problem.
\newblock In {\em 2015 Conference on technologies and applications of
  artificial intelligence (TAAI)}, pages 136--141. IEEE, 2015.

\bibitem{browne2012survey}
Cameron~B Browne, Edward Powley, Daniel Whitehouse, Simon~M Lucas, Peter~I
  Cowling, Philipp Rohlfshagen, Stephen Tavener, Diego Perez, Spyridon
  Samothrakis, and Simon Colton.
\newblock A survey of monte carlo tree search methods.
\newblock {\em IEEE Transactions on Computational Intelligence and AI in
  games}, 4(1):1--43, 2012.

\bibitem{reijnen2023job}
Robbert Reijnen, Kjell van Straaten, Zaharah Bukhsh, and Yingqian Zhang.
\newblock Job shop scheduling benchmark: Environments and instances for
  learning and non-learning methods.
\newblock {\em arXiv preprint arXiv:2308.12794}, 2023.

\bibitem{demirkol1998benchmarks}
Ebru Demirkol, Sanjay Mehta, and Reha Uzsoy.
\newblock Benchmarks for shop scheduling problems.
\newblock {\em European Journal of Operational Research}, 109(1):137--141,
  1998.

\bibitem{cpsatlp}
Laurent Perron and Frédéric Didier.
\newblock Cp-sat.

\end{thebibliography}

\end{document}